\documentclass{article} 

\usepackage{microtype}
\usepackage{graphicx}
\usepackage{subfigure}
\usepackage{booktabs} 

\usepackage{hyperref}

\usepackage[accepted]{icml2025_pral}

\usepackage{amsmath}
\usepackage{amssymb}
\usepackage{mathtools}
\usepackage{amsthm}

\usepackage[capitalize,noabbrev]{cleveref}

\usepackage[textsize=tiny]{todonotes}

\icmltitlerunning{Representing Prompting Patterns with PDL: Compliance Agent Case Study}

\usepackage{booktabs}   \usepackage{subcaption} 

\usepackage{alltt}
\usepackage{graphicx}
\usepackage{listings}
\usepackage{mathpartir}

\definecolor{keyword}{HTML}{37AC4A} 
\definecolor{jinja2}{HTML}{0070C0}
\definecolor{comment}{HTML}{7F7F7F}
\definecolor{generated}{HTML}{37AC4A} 
\definecolor{toolout}{HTML}{A51DFF}

\lstdefinelanguage{pdl}{
  xleftmargin=4mm,
  numbers=left,
  numbersep=2mm,
  numberstyle=\tiny\color{gray},
  basicstyle=\fontsize{8}{9}\ttfamily,
  basewidth=0.5em,
  alsoletter={:},
  morekeywords={args:,array:,as:,call:,case:,code:,content:,contribute:,data:,def:,defs:,description:,else:,for:,function:,if:,include:,input:,join:,lang:,lastOf:,match:,message:,model:,multiline:,num_iterations:,object:,parameters:,parser:,pdl_context:,read:,repeat:,return:,role:,spec:,text:,then:,type:,until:,with:},
  keywordstyle=\color{keyword}\bf,
  morestring=[s]{$\{}{\}},
  stringstyle=\color{jinja2},
  emphstyle=\color{jinja2},
  showstringspaces=false,
  morecomment=[l]{\#},
  commentstyle=\color{comment}\it,
}

\begin{document}

\twocolumn[
\icmltitle{Representing Prompting Patterns with PDL: Compliance Agent Case Study}

\begin{icmlauthorlist}
\icmlauthor{Mandana Vaziri}{ykt}
\icmlauthor{Louis Mandel}{ykt}
\icmlauthor{Yuji Watanabe}{trl}
\icmlauthor{Hirokuni Kitahara}{trl}
\icmlauthor{Martin Hirzel}{ykt}
\icmlauthor{Anca Sailer}{ykt}
\end{icmlauthorlist}

\icmlaffiliation{ykt}{IBM Research, Yorktown Heights, USA}
\icmlaffiliation{trl}{IBM Research, Tokyo, Japan}

\icmlcorrespondingauthor{Mandana Vaziri}{mvaziri@us.ibm.com}

\icmlkeywords{Prompt Programming, Agents}

\vskip 0.3in
]

\printAffiliationsAndNotice{}  

\begin{abstract}

Prompt engineering for LLMs remains complex, with existing frameworks either hiding complexity behind restrictive APIs or providing inflexible canned patterns that resist customization -- making sophisticated agentic programming challenging. We present the Prompt Declaration Language~(PDL), a novel approach to prompt representation that tackles this fundamental complexity by bringing prompts to the forefront, enabling manual and automatic prompt tuning while capturing the composition of LLM calls together with rule-based code and external tools. By abstracting away the plumbing for such compositions, PDL aims at improving programmer productivity while providing a declarative representation that is amenable to optimization. This paper demonstrates PDL's utility through a real-world case study of a compliance agent. Tuning the prompting pattern of this agent yielded up to 4x performance improvement compared to using a canned agent and prompt pattern.

\end{abstract}

\section{Introduction}
\label{intro}
Prompt engineering for large language models (LLMs) has been notoriously difficult.
Small prompt variations have an outsized impact on the results, prompts are model dependent, and prompting patterns are published informally. Recent years have seen the rise of a variety of prompt programming languages and frameworks.
Low-level prompt languages, such as Guidance~\cite{microsoft_2023}, LMQL~\cite{beurerkellner_fischer_vechev_2023}, and SGLang~\cite{zheng_et_al_2023}, give developers exact control over the prompts to express multi-turn interactions with LLMs, and offer additional benefits such as constrained decoding to help shape the output of LLMs, and runtime performance optimizations (parallelism, KV prefix caching).
However, being low-level makes them ill-suited as a declarative representation.
High-level prompt frameworks such as LangChain/LangGraph~\cite{chase_et_al_2022}, LLama Stack~\cite{llamastack}, and others provide APIs encapsulating various prompting patterns, such as CoT~\cite{wei_et_al_2022-prompting} and ReAct~\cite{yao_et_al_2023} or ReWoo~\cite{xu_et_al_2023} that form the basis of agent development. Agentic frameworks such as AutoGen~\cite{wu_et_al_2023} and Crew\-AI~\cite{moura_2023} have adopted the concept of agent as the main organizing feature and have focused on the ReAct agentic prompting pattern and its variations.
Most of these approaches bury prompts in imperative code or behind APIs, making prompting patterns hard to customize.
However, in practice, prompting patterns need to be customized to implement AI agents successfully and to maintain them as the underlying LLMs evolve~\cite{schluntz_zhang_2024}.

The Prompt Declaration Language (PDL) is a programming language for specifying LLM prompts and LLM-based workflows and agents~\cite{vaziri_et_al_2024}.
At its core, PDL is a declarative representation, written in YAML, and captures the composition of model calls together with rule-based traditional code. PDL brings prompts to the forefront and abstracts away the plumbing necessary for such compositions. It provides a set of orthogonal language features allowing developers to express their own prompting patterns, and aims at improving programmer productivity. As LLMs have evolved, their interface is no longer string in and string out. Instead, their input is a structured list of {\em messages}, consisting of {\em role} and {\em content} that capture a history of multi-turn LLM interactions and tool calling. The PDL interpreter accumulates such messages implicitly and hides their underlying structure
to free the developer to think at a higher level of abstraction. PDL is a typed language using JSON Schema~\cite{pezoa_et_al_2016} as types, and can type-check both the input and output of models. PDL types are seamlessly integrated with constrained decoding~\cite{willard_louf_2023} in platforms and models that support it, to ensure the shape of the output. PDL leverages LiteLLM~\cite{litellm_2023} to support a wide variety of models and model providers. It also handles {\em chat APIs} -- the specific formatting of structured messages into strings -- seamlessly across models, making it easier to adapt a program to use different models.

This paper first gives an overview of the PDL representation for prompting patterns (Sec.~\ref{overview}). It then presents a real-world case study demonstrating the expressivity and usefulness of PDL as a language for developing agents (Sec.~\ref{studies}). The case study uses PDL in an agent for CISO IT compliance task automation. It demonstrates a significant~(up to 4x) performance improvement across a series of models when using PDL compared to an architecture that does not.
 
\section{Introduction to PDL}
\label{overview}

At the heart of agents~\cite{yao_et_al_2023} is the ability to decide
when to use a tool, which one, and how. Figure~\ref{tool}(a) shows a simple PDL program that uses one tool. The program is written in YAML, which makes it easy to see properly formatted prompts. PDL adds enough scripting to allow users to include not only their textual prompts in YAML, but also entire prompting patterns. Lines~2 to 17 contain definitions, in this case defining the tool (Wikipedia search) and assigning it to variable \lstinline{search}.
Starting at line~18, we see the {\em block}s that constitute this program. In PDL, the program computes a result while also implicitly maintaining and updating a background {\em context} of {\em messages}. This context gets used as input when making LLM calls. So each block has a result, and also contributes that result to the background context. Line~18 starts a \lstinline{text} block: it takes each block in the list, stringifies its result, and concatenates them as the result of the \lstinline{text}. Alternatively, an \lstinline{array} block could be used to generate an array of results instead; \lstinline{lastOf} acts like a sequence and returns the result of the last block (not shown in this figure). Line~19 specifies the {\em system} prompt with a {\em message} block (indicated by the \lstinline{content} field). 
Notice how YAML renders the longer prompt in a natural way, making it more readable than if it had been buried in imperative code. Line~26 defines a {\em tools} prompt, indicating the tools available to the model. Line~28 uses a Jinja expression to access the value of the attribute \lstinline{signature} of the function \lstinline{search} that contains the signature of the tool extracted from the function definition. Line~30 contains the query we wish to send to the LLM. 

\begin{figure}[ht!]
\begin{minipage}[b]{0.52\textwidth}
\begin{lstlisting}[emph={actions,arguments:,name:,search:,topic:}]
description: tool use
defs:
  search:
    description: Wikipedia search
    function:
      topic:
        type: string
        description: Topic to search
    return:
      lang: python
      code: |
        import warnings, wikipedia
        warnings.simplefilter("ignore")
        try:
          result = wikipedia.summary("${ topic }")
        except wikipedia.WikipediaException as e:
          result = str(e)
text:
- role: system
  content: >
    You are a helpful AI assistant with access to the 
    following tools. If a tool does not exist in the 
    provided list of tools, notify the user that you 
    do not have the ability to fulfill the request.
  contribute: [context]
- role: tools
  content:
    text: ${ [ search.signature ] }
  contribute: [context]
- "What is the circumference of planet Earth?\n"
- def: actions
  model: ollama_chat/granite3.3:8b
  parser: json
  spec: [{ name: str, arguments: { topic: str }}]
- "\n"
- if: ${ actions[0].name == "search" }
  then:
    call: ${ search }
    args:
      topic: ${ actions[0].arguments.topic }
\end{lstlisting}
\centerline{(a) Code}
\end{minipage}
\hfill
\begin{minipage}[b]{0.47\textwidth}
{\small\vspace{2mm}
\textcolor{generated}{[\{'name': 'search', 'arguments': \{'topic': 'circumference of Earth'\}\}]}

\textcolor{toolout}{Earth's circumference is the distance around Earth. Measured around the equator, it is 40,075.017 km (24,901.461 mi). Measured passing through the poles, the circumference is 40,007.863 km (24,859.734 mi).} }\\[2mm]
\centerline{(b) Interpreter trace}
\end{minipage}
\caption{\label{tool}Simple Tool Use in PDL}
\end{figure}

Lines~31 to 34 show a model call, in this case, to a local Ollama model, \lstinline{granite3.3:8b}. PDL supports a wide variety of models and model providers because it uses LiteLLM as its backend.
Line~31 defines variable \lstinline{actions} to contain the result of the model call.
The input to this model call is the context accumulated from executing the blocks so far (system prompt, tools prompt, and user query).
This block could also contain any parameters we wish to send to the model. Line~33 indicates that the output of the model should be parsed as JSON, and line~34 specifies a type for the output: a list of objects with two attributes, \lstinline{name} and \lstinline{arguments}. The PDL interpreter automatically checks the output against this type, and also uses the type to set up appropriate parameters for constrained decoding on various platforms, to make the LLM produce output of this shape.

Line~36 is a conditional. Although PDL is a YAML-based representation, it supports control structures such as conditionals and loops. It also supports modularity and reuse through function definitions and importing PDL code from other files or libraries. This design choice enables entire prompting patterns to be expressed in YAML, as opposed to being split apart into YAML and, for example, Python, as is typically the case. The if-statement checks to see if the action returned by the LLM is a search, in which case it calls the function \lstinline{search}. 
The body of the function~(defined in Lines~10--17) uses a Python code block to perform the Wikipedia search. 
PDL supports different kinds of code blocks and allows the composition of LLMs and code, abstracting away all the plumbing necessary for such compositions. It is a representation that allows users to see prompts in the forefront while developing prompting patterns and agents. Because it is a higher-level representation, it is also a good target for automated optimization. \citet{spiess_et_al_2024} used PDL as a generation target for automated prompt pattern search.

Figure~\ref{tool}(b) shows the output of the interpreter when this program is executed. The output of the LLM call is shown in green. The LLM chooses to use a tool and specifies the arguments for the tool. The output of the code block is shown in purple and is the output of the Wikipedia call.

\section{Case Study}
\label{studies}
We used a real-world application for our case study: Chief Information Security Officer (CISO) Compliance Agent, which is an AI agent for automating IT compliance tasks.

Compliance tasks traditionally demand specialized expertise due to their reliance on complex standards and internal organizational policies. While automation tools exist for routine operations, policy assessment remains manual at large as the compliance teams are mostly non-technical. Our CISO Agent addresses this gap by combining LLM reasoning with the ReAct pattern~\cite{yao_et_al_2023} to provide automated programmatic development support to those teams. Upon receiving new regulatory requirements, the agent comprehends their content, identifies the target systems, generates and deploys necessary scripts, validates their outcomes, and provides a comprehensive posture reporting.

\begin{figure}[htbp]
  \centering

  \begin{minipage}[b]{0.48\columnwidth}
    \includegraphics[width=\linewidth]{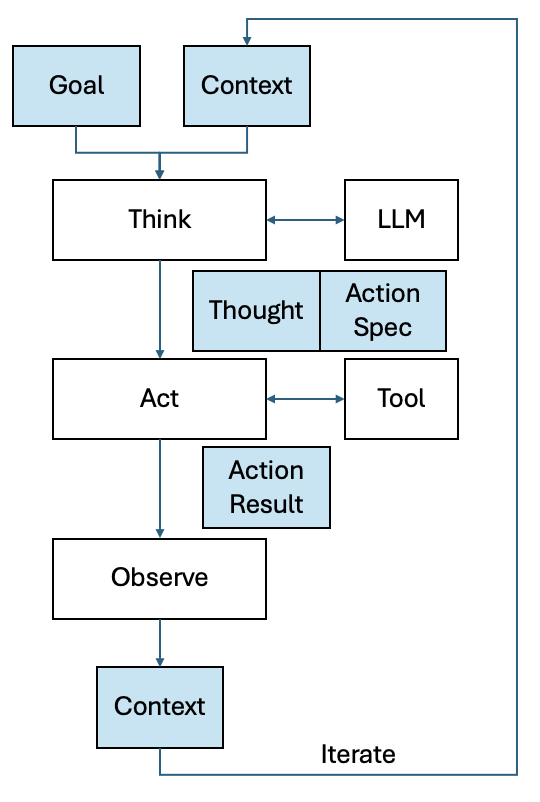}
    \caption*{(a) Original ReAct}
  \end{minipage}
  \hfill
  \begin{minipage}[b]{0.48\columnwidth}
    \includegraphics[width=\linewidth]{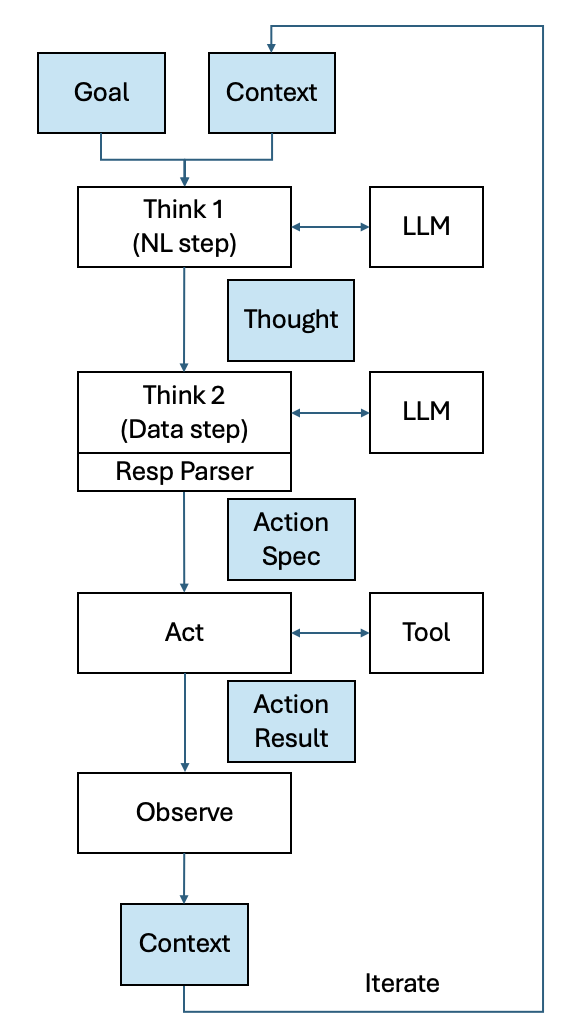}
    \caption*{(b) PDL-based}
  \end{minipage}

  \caption{CISO Agent Architecture}
  \label{fig:ciso_agent_architecture}
\end{figure}

The first version of our CISO Agent was implemented using a traditional ReAct pattern with CrewAI \cite{moura_2023}. We then experimented using PDL to introduce pattern and prompts customization particularly needed with compact, more affordable LLMs. Figure~\ref{fig:ciso_agent_architecture} shows the two architectures: (a)~original ReAct pattern implementation, and (b)~PDL-based architecture optimized for compact LLMs.

In the original ReAct pattern, the task description is the first input (Goal in Figure~\ref{fig:ciso_agent_architecture}(a)). In the Think step, the LLM receives the Goal and considers the best next action. This step has two outputs: a natural language text for the next action (Thought), and a tool to call in JSON format with its parameters (ActionSpec). Then, in the Act step, the specified tool is executed based on this ActionSpec data. The result of this execution (ActionResult) is fed back into Context in the Observe step. This updated Context is used as another input for the next Think step.
However, this pattern often does not work well with smaller LLMs. A typical failure example is that outputting Thought in natural language and ActionSpec as a JSON string at the same time causes syntax errors in JSON output resulting in tool call failures or hallucination of tool names in the ActionSpec.

As a solution to these problems, the CISO Agent developers devised a PDL-based agent architecture shown in Figure~\ref{fig:ciso_agent_architecture}(b).

First, since small models tend to produce corrupted ActionSpec if Thought and ActionSpec are output simultaneously, this new architecture splits Think into two stages, Think1 (natural language step) which outputs Thought, and Think2~(data step) which outputs ActionSpec. Traditional agent frameworks such as CrewAI lack the customization capabilities that PDL provides for core agent workflow modification—a crucial differentiator.

Furthermore, even with the two-stage design, Think2 exhibits model-specific failures, for example, producing \mbox{\texttt{\detokenize{{"name": "abc"}}}} while the expected format is \texttt{\detokenize{{"tool_name": "abc"}}}. Our PDL-based agent addresses these issues through a custom Response Parser that correctly handles ActionSpec outputs, thus demonstrating PDL's flexibility, practical benefits, and its capabilities beyond the traditional frameworks scope.

\begin{figure}[htbp]
  \centering
  \includegraphics[width=0.5\textwidth]{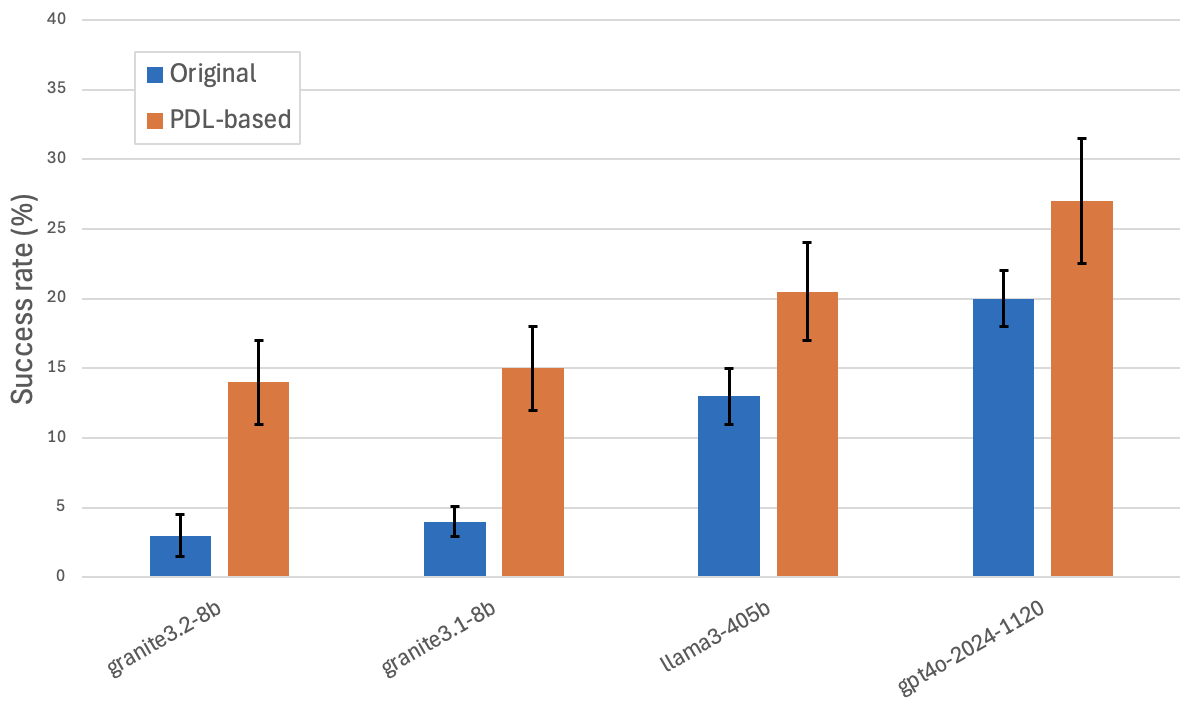}
  \caption{CISO Agent Evaluation for Original (blue) and PDL-based (orange) Architecture.}
  \label{fig:ciso_comparison}
\end{figure}

Figure~\ref{fig:ciso_comparison} presents performance evaluation results using ITBench~\cite{itbench_paper}, comparing the original and PDL-based CISO Agent implementations. Both versions use identical models and tools, differing only in agent architecture. The PDL implementation demonstrates consistent improvements across all models, with particularly dramatic gains in smaller models like granite3.2-8b, achieving 4 times better performance.

\begin{figure}[!t]
  \centering
  \begin{tabular}{p{15mm}l}
  (a) \mbox{Original} (modelA) &
  \begin{minipage}{0.61\columnwidth}
    \includegraphics[width=\linewidth]{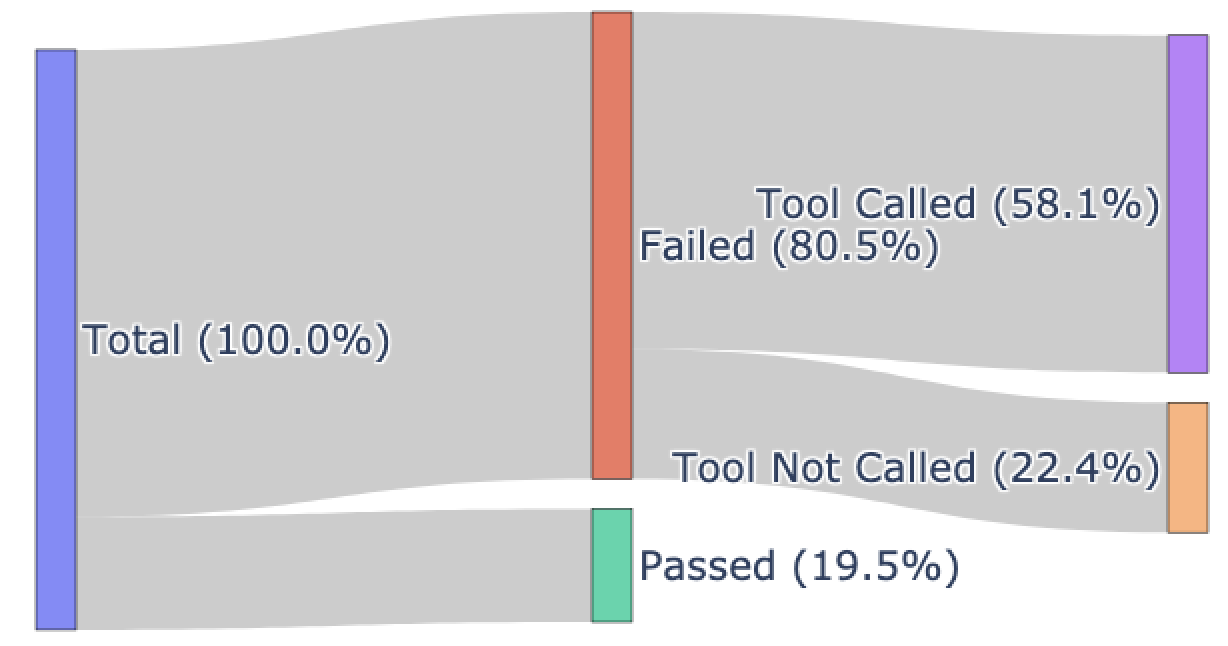}
  \end{minipage}\\
  (b) \mbox{PDL-based} (modelA) &
  \begin{minipage}{0.61\columnwidth}
    \includegraphics[width=\linewidth]{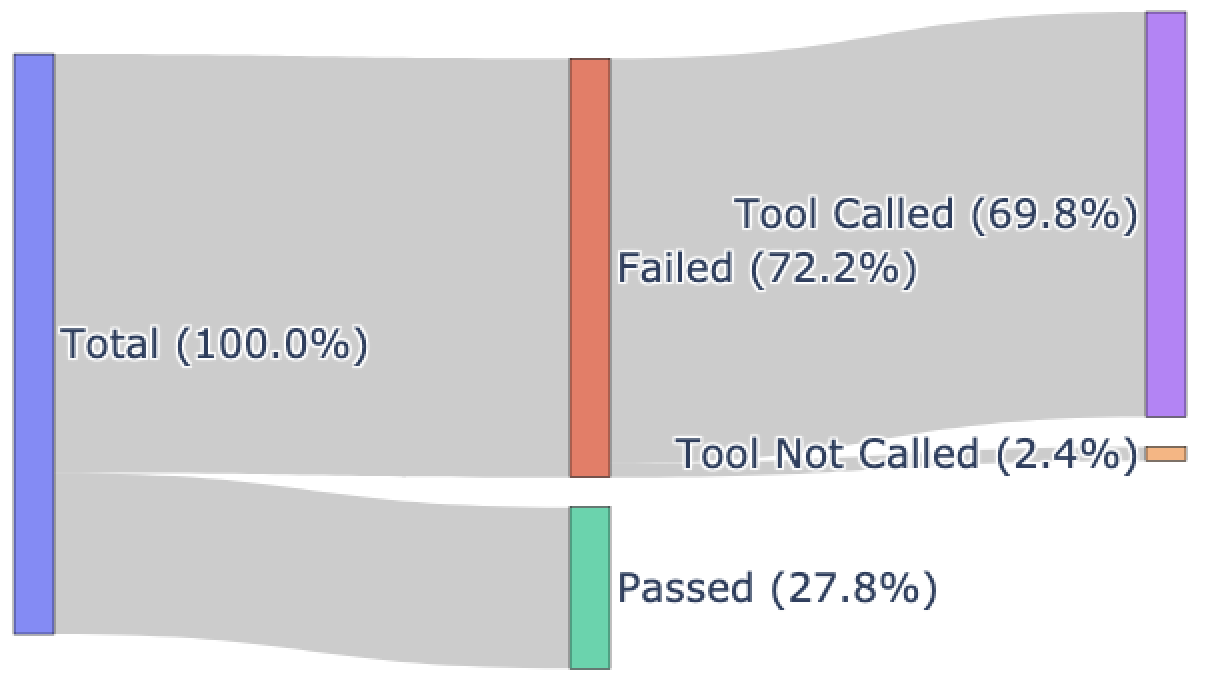}
  \end{minipage}\\
  (c) \mbox{Original} (modelB) &
  \begin{minipage}{0.61\columnwidth}
    \includegraphics[width=\linewidth]{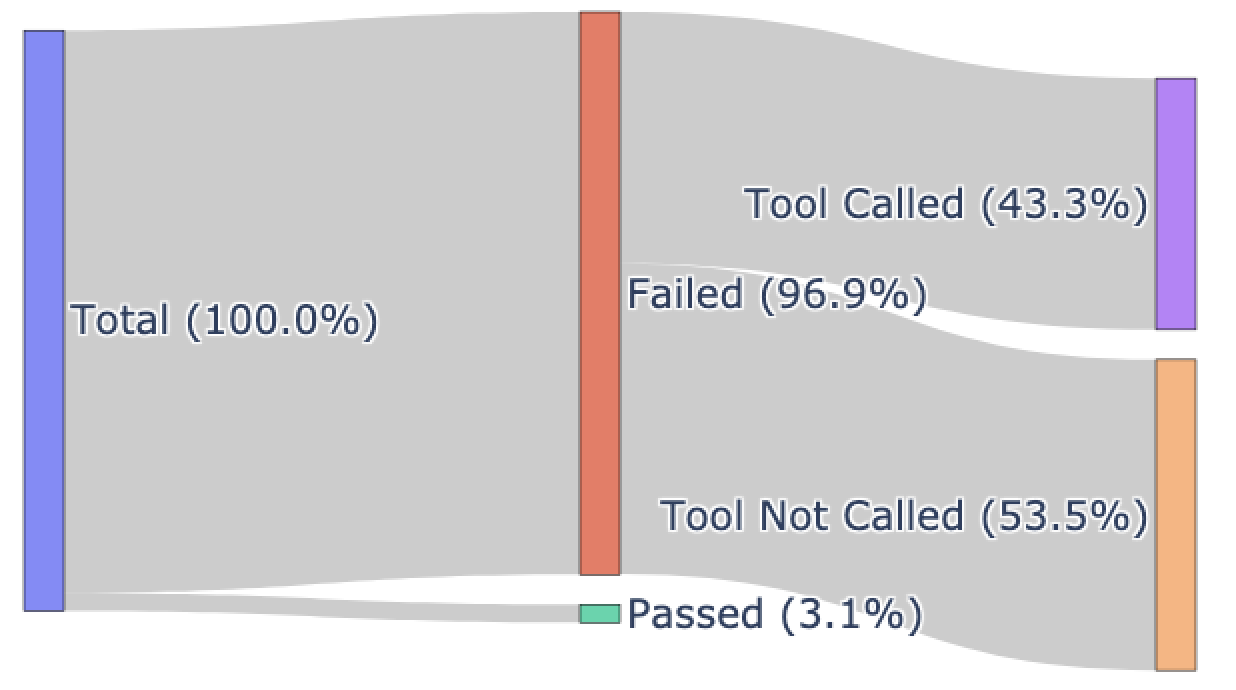}
  \end{minipage}\\
  (d) \mbox{PDL-based} (modelB) &
  \begin{minipage}{0.61\columnwidth}
    \includegraphics[width=\linewidth]{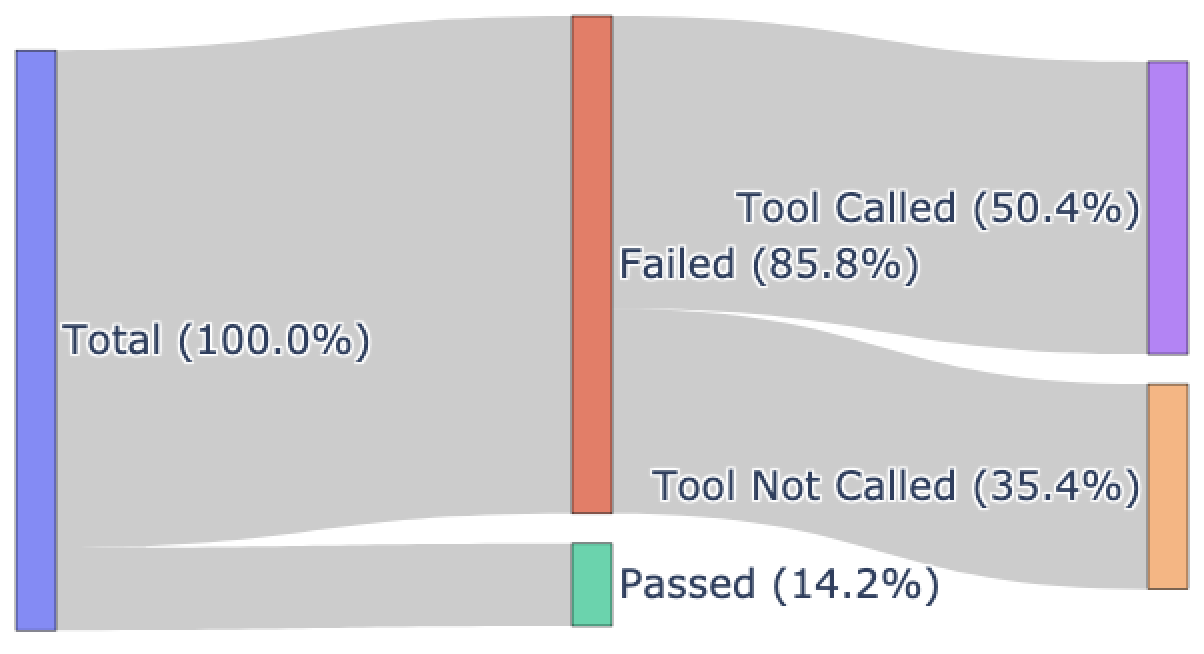}
  \end{minipage}
  \end{tabular}
  \caption{Tool Call Success Rate Comparison (modelA: gpt4o-2024-11-20, modelB: granite3.2-8b-instruct)}
  \label{fig:tool_call_analysis}
\end{figure}

Performance analysis reveals that PDL's improvements stem primarily from reduced tool call failures. Figure~\ref{fig:tool_call_analysis} displays four Sankey diagrams representing approximately 200 evaluation tests per condition, illustrating the relationship between test success and tool call execution accuracy. The first two diagrams (a) and (b) show results when using gpt4o-2024-11-20, (a) corresponding to the original CrewAI implementation, and (b) to the PDL-based implementation. The important difference here is that the cases where the tool was not called dropped from 22.4\% to 2.4\% in the PDL-based implementation, leading to a significant increase in overall task success rates.

Diagrams (c) and (d) show results when using granite3.2-8b-instruct as the LLM. Here, cases where no tool was called decreased from 53.5\% to 35.4\%, accounting for the 4 times improvement mentioned above in the success rate.

These improvements result from PDL's extensive LLM interaction customization capabilities. The findings highlight how PDL's fine-grained control mechanisms prove essential for optimizing AI agent performance, especially when using smaller, resource-constrained language models.

\section{Related Work}
\label{related}
PDL is a domain-specific language (DSL): a program representation that
aims to be both easier to use (for programmers) and easier to transform
(for code generators or optimizers)~\cite{mernik_heering_sloane_2005}
than general-purpose programming languages such as Python.
While PDL is not the only DSL for prompting, other such DSLs such as
LMQL~\cite{beurerkellner_fischer_vechev_2023} or
DSPy~\cite{zheng_et_al_2023} are embedded in Python.
In contrast, PDL is embedded in the YAML data representation format,
making it easier to manipulate programmatically than Python's rich
imperative syntax.
Prompt optimizers such as DSPy~\cite{khattab_et_al_2024} or
Evo\-Agent~\cite{yuan_et_al_2024} rewrite prompts to improve their
predictive accuracy on a given dataset and task.
Unfortunately, unlike PDL, they generate a representation that is
ill-suited for programmers to read, let alone tweak further.
Agent frameworks like Auto\-Gen~\cite{wu_et_al_2023} and
CrewAI~\cite{moura_2023} have built-in prompts for agents buried
and scattered around the framework implementation, making prompts
difficult to adapt for novel tasks or models.
In contrast, PDL keeps prompts at the forefront, with a unified
representation that encompasses prompts along with declarative
agentic logic.

\section{Conclusion}
\label{conclusion}

This paper introduced PDL, a novel prompt representation that prioritizes prompt visibility while enabling seamless composition of LLM calls with rule-based code. The real-world case study demonstrated significant performance improvements through PDL implementation. 
In the future, we will explore PDL as a target of LLM generation to show the versatility of the representation as a way for developers to express their prompting patterns, as well as for LLMs to generate plans of action.

\section*{Impact Statement}

This paper presents research focused on advancing prompting techniques for large language models. The work carries potential societal implications that fall within the general scope of LLM development and deployment,  none of which requires specific emphasis in this context.

\bibliography{bibfile}

\begin{thebibliography}{20}
\providecommand{\natexlab}[1]{#1}
\providecommand{\url}[1]{\texttt{#1}}
\expandafter\ifx\csname urlstyle\endcsname\relax
  \providecommand{\doi}[1]{doi: #1}\else
  \providecommand{\doi}{doi: \begingroup \urlstyle{rm}\Url}\fi

\bibitem[BerryAI(2025)]{litellm_2023}
BerryAI.
\newblock {LiteLLM}, July 2025.
\newblock URL \url{https://docs.litellm.ai/}.

\bibitem[Beurer-Kellner et~al.(2023)Beurer-Kellner, Fischer, and
  Vechev]{beurerkellner_fischer_vechev_2023}
Beurer-Kellner, L., Fischer, M., and Vechev, M.
\newblock Prompting is programming: A query language for large language models.
\newblock In \emph{Conference on Programming Language Design and Implementation
  (PLDI)}, pp.\  1946--1969, June 2023.

\bibitem[{Chase et al.}(2025)]{chase_et_al_2022}
{Chase et al.}, H.
\newblock {LangChain}, July 2025.
\newblock URL \url{https://github.com/langchain-ai/langchain}.

\bibitem[Jha et~al.(2025)Jha, Arora, Watanabe, Yanagawa, Chen, Clark, Bhavya,
  Verma, Kumar, Kitahara, Zheutlin, Takano, Pathak, George, Wu, Turkkan,
  Vanloo, Nidd, Dai, Chatterjee, Gupta, Samanta, Aggarwal, Lee, Murali, wook
  Ahn, Kar, Rahane, Fonseca, Paradkar, Deng, Moogi, Mohapatra, Abe,
  Narayanaswami, Xu, Varshney, Mahindru, Sailer, Shwartz, Sow, Fuller, and
  Puri]{itbench_paper}
Jha, S., Arora, R., Watanabe, Y., Yanagawa, T., Chen, Y., Clark, J., Bhavya,
  B., Verma, M., Kumar, H., Kitahara, H., Zheutlin, N., Takano, S., Pathak, D.,
  George, F., Wu, X., Turkkan, B.~O., Vanloo, G., Nidd, M., Dai, T.,
  Chatterjee, O., Gupta, P., Samanta, S., Aggarwal, P., Lee, R., Murali, P.,
  wook Ahn, J., Kar, D., Rahane, A., Fonseca, C., Paradkar, A., Deng, Y.,
  Moogi, P., Mohapatra, P., Abe, N., Narayanaswami, C., Xu, T., Varshney,
  L.~R., Mahindru, R., Sailer, A., Shwartz, L., Sow, D., Fuller, N. C.~M., and
  Puri, R.
\newblock {ITBench}: Evaluating {AI} agents across diverse real-world {IT}
  automation tasks.
\newblock In \emph{International Conference on Machine Learning (ICML)}, June
  2025.

\bibitem[Khattab et~al.(2024)Khattab, Singhvi, Maheshwari, Zhang, Santhanam, A,
  Haq, Sharma, Joshi, Moazam, Miller, Zaharia, and Potts]{khattab_et_al_2024}
Khattab, O., Singhvi, A., Maheshwari, P., Zhang, Z., Santhanam, K., A, S.~V.,
  Haq, S., Sharma, A., Joshi, T.~T., Moazam, H., Miller, H., Zaharia, M., and
  Potts, C.
\newblock {DSPy}: Compiling declarative language model calls into
  self-improving pipelines.
\newblock In \emph{International Conference on Learning Representations
  (ICLR)}, May 2024.

\bibitem[Mernik et~al.(2005)Mernik, Heering, and
  Sloane]{mernik_heering_sloane_2005}
Mernik, M., Heering, J., and Sloane, A.~M.
\newblock When and how to develop domain-specific languages.
\newblock \emph{ACM Computing Surveys (CSUR)}, 37\penalty0 (4):\penalty0
  316--344, 2005.

\bibitem[Meta(2025)]{llamastack}
Meta.
\newblock Llama {Stack}, July 2025.
\newblock URL \url{https://github.com/meta-llama/llama-stack}.

\bibitem[Microsoft(2025)]{microsoft_2023}
Microsoft.
\newblock \{guidance\}: A guidance language for controlling large language
  models, July 2025.
\newblock URL \url{https://github.com/guidance-ai/guidance}.

\bibitem[Moura(2025)]{moura_2023}
Moura, J.
\newblock {CrewAI}: Framework for orchestrating role-playing, autonomous {AI}
  agents, July 2025.
\newblock URL \url{https://github.com/crewAIInc/crewAI}.

\bibitem[Pezoa et~al.(2016)Pezoa, Reutter, Suarez, Ugarte, and
  Vrgo\v{c}]{pezoa_et_al_2016}
Pezoa, F., Reutter, J.~L., Suarez, F., Ugarte, M., and Vrgo\v{c}, D.
\newblock Foundations of {JSON} schema.
\newblock In \emph{International Conference on World Wide Web (WWW)}, pp.\
  263--273, April 2016.

\bibitem[Schluntz \& Zhang(2025)Schluntz and Zhang]{schluntz_zhang_2024}
Schluntz, E. and Zhang, B.
\newblock Building effective agents, July 2025.
\newblock URL
  \url{https://www.anthropic.com/research/building-effective-agents}.

\bibitem[Spiess et~al.(2025)Spiess, Vaziri, Mandel, and
  Hirzel]{spiess_et_al_2024}
Spiess, C., Vaziri, M., Mandel, L., and Hirzel, M.
\newblock {AutoPDL}: Automatic prompt optimization for {LLM} agents.
\newblock In \emph{Conference on Automated Machine Learning (AutoML)},
  September 2025.

\bibitem[Vaziri et~al.(2024)Vaziri, Mandel, Spiess, and
  Hirzel]{vaziri_et_al_2024}
Vaziri, M., Mandel, L., Spiess, C., and Hirzel, M.
\newblock {PDL}: A declarative prompt programming language, October 2024.
\newblock URL \url{http://arxiv.org/abs/2410.19135}.

\bibitem[Wei et~al.(2022)Wei, Wang, Schuurmans, Bosma, Ichter, Xia, Chi, Le,
  and Zhou]{wei_et_al_2022-prompting}
Wei, J., Wang, X., Schuurmans, D., Bosma, M., Ichter, B., Xia, F., Chi, E., Le,
  Q., and Zhou, D.
\newblock Chain-of-thought prompting elicits reasoning in large language
  models.
\newblock In \emph{Advances in Neural Information Processing Systems
  (NeurIPS)}, pp.\  24824--24837, December 2022.

\bibitem[Willard \& Louf(2023)Willard and Louf]{willard_louf_2023}
Willard, B.~T. and Louf, R.
\newblock Efficient guided generation for large language models, July 2023.
\newblock URL \url{https://arxiv.org/abs/2307.09702}.

\bibitem[Wu et~al.(2023)Wu, Bansal, Zhang, Wu, Li, Zhu, Jiang, Zhang, Zhang,
  Liu, Awadallah, White, Burger, and Wang]{wu_et_al_2023}
Wu, Q., Bansal, G., Zhang, J., Wu, Y., Li, B., Zhu, E., Jiang, L., Zhang, X.,
  Zhang, S., Liu, J., Awadallah, A.~H., White, R.~W., Burger, D., and Wang, C.
\newblock {AutoGen}: Enabling next-gen {LLM} applications via multi-agent
  conversation, October 2023.
\newblock URL \url{https://arxiv.org/abs/2308.08155}.

\bibitem[Xu et~al.(2023)Xu, Peng, Lei, Mukherjee, and Xu]{xu_et_al_2023}
Xu, B., Peng, Z., Lei, B., Mukherjee, S., and Xu, D.
\newblock Decoupling reasoning from observations for efficient augmented
  language models, September 2023.
\newblock URL \url{https://openreview.net/forum?id=CpgoO6j6W1}.

\bibitem[Yao et~al.(2023)Yao, Zhao, Yu, Du, Shafran, Narasimhan, and
  Cao]{yao_et_al_2023}
Yao, S., Zhao, J., Yu, D., Du, N., Shafran, I., Narasimhan, K.~R., and Cao, Y.
\newblock {ReAct}: Synergizing reasoning and acting in language models.
\newblock In \emph{International Conference on Learning Representations
  (ICLR)}, May 2023.

\bibitem[Yuan et~al.(2024)Yuan, Song, Chen, Tan, Li, and Yang]{yuan_et_al_2024}
Yuan, S., Song, K., Chen, J., Tan, X., Li, D., and Yang, D.
\newblock {EvoAgent}: Towards automatic multi-agent generation via evolutionary
  algorithms, July 2024.
\newblock URL \url{https://arxiv.org/abs/2406.14228}.

\bibitem[Zheng et~al.(2023)Zheng, Yin, Xie, Huang, Sun, Yu, Cao, Kozyrakis,
  Stoica, Gonzalez, Barrett, and Sheng]{zheng_et_al_2023}
Zheng, L., Yin, L., Xie, Z., Huang, J., Sun, C., Yu, C.~H., Cao, S., Kozyrakis,
  C., Stoica, I., Gonzalez, J.~E., Barrett, C., and Sheng, Y.
\newblock Efficiently programming large language models using {SGLang},
  December 2023.
\newblock URL \url{https://arxiv.org/abs/2312.07104}.

\end{thebibliography}
\bibliographystyle{icml2025}

\end{document}